\documentclass[12pt]{iopart}

\usepackage{graphicx}
\usepackage{subfigure}
\usepackage{cite}
\usepackage[export]{adjustbox}
\usepackage{xcolor}
\usepackage{hyperref}
\usepackage{multirow}
\usepackage{multicol}
\usepackage{xcolor}
 
\begin{document}

\title[Author guidelines for IOP Publishing journals in  \LaTeXe]{Assessing visual acuity in visual prostheses through a virtual-reality system}

\author{Melani Sanchez-Garcia\textsuperscript{*}$^1$, Roberto Morollon-Ruiz\textsuperscript{*}$^2$, Ruben Martinez-Cantin$^3$, Jose J. Guerrero$^3$, Eduardo Fernandez-Jover$^2$}

\address{$^1$Department of Computer Science. University of California, Santa Barbara, CA, USA}
\address{$^2$Instituto de Bioingeniería. Universidad de Miguel Hernandez, Spain}
\address{$^3$Instituto de Investigación en Ingeniería de Aragón, (I3A). Universidad de Zaragoza, Spain}
\ead{mesangar@ucsb.edu, rmorollon@umh.es}

\vspace{10pt}

\begin{abstract} 
\textit{Objective.} Current visual implants still provide very low resolution and limited field of view, thus limiting visual acuity in implanted patients. Developments of new strategies of artificial vision simulation systems by harnessing new advancements in technologies are of upmost priorities for the development of new visual devices. \textit{Approach}. In this work, we take advantage of virtual-reality software paired with a portable head-mounted display and evaluated the performance of normally sighted participants under simulated prosthetic vision with variable field of view and number of pixels. Our simulated prosthetic vision system allows simple experimentation in order to study the design parameters of future visual prostheses. Ten normally sighted participants volunteered for a visual acuity study. Subjects were required to identify computer-generated Landolt-C gap orientation and different stimulus based on light perception, time-resolution, light location and motion perception commonly used for visual acuity examination in the sighted. Visual acuity scores were recorded across different conditions of number of electrodes and size of field of view. \textit{Main results.} Our results showed that of all conditions tested, a field of view of ${20^\circ}$ and 1000 phosphenes of resolution proved the best, with a visual acuity of 1.3 logMAR. Furthermore, performance appears to be correlated with phosphene density, but showing a diminishing return when field of view is less than $20^\circ$. \textit{Significance}. The development of new artificial vision simulation systems can be useful to guide the development of new visual devices and the optimization of field of view and resolution to provide a helpful and valuable visual aid to profoundly or totally blind patients.

\end{abstract}
\vspace{2pc}


\section{Introduction}


Low vision or blindness are major health issues for the individual’s quality of life. The leading causes of blindness are primarily age-related eye diseases such as age-related macular degeneration (AMD) \cite{vannewkirk2000prevalence,vingerling1995prevalence}, cataract, retinitis pigmentosa (RP)\cite{hartong2006retinitis} and glaucoma \cite{steinmetz2021causes}. The loss of photoreceptors due to degeneration is a major cause of vision loss, resulting in dysfunctional light detection, transduction, and transmission \cite{curcio2000spare,busskamp2010genetic,yue2016retinal}. In the case of RP, these inherited disorders can affect either rods or cone primarily. The most common form of RP is characterized initially by night blindness, followed by progressive loss in the peripheral field of view in daylight, eventually leading to blindness after several decades. The case of AMD is characterized by sudden acuity loss. Currently, there is no cure for RP or AMD \cite{richer2004double,heier2012intravitreal}. However, great efforts have been devoted to restoring the resulting poor visual function, based primarily on gene therapy, stem cell transplantation, or visual prosthesis \cite{stieger2010preclinical,huang2011stem,wong2011promises,beltran2012gene,takahashi2018gene,bloch2019advances}.

\begin{figure}[t!]
\centering
{\includegraphics[width = 5in]{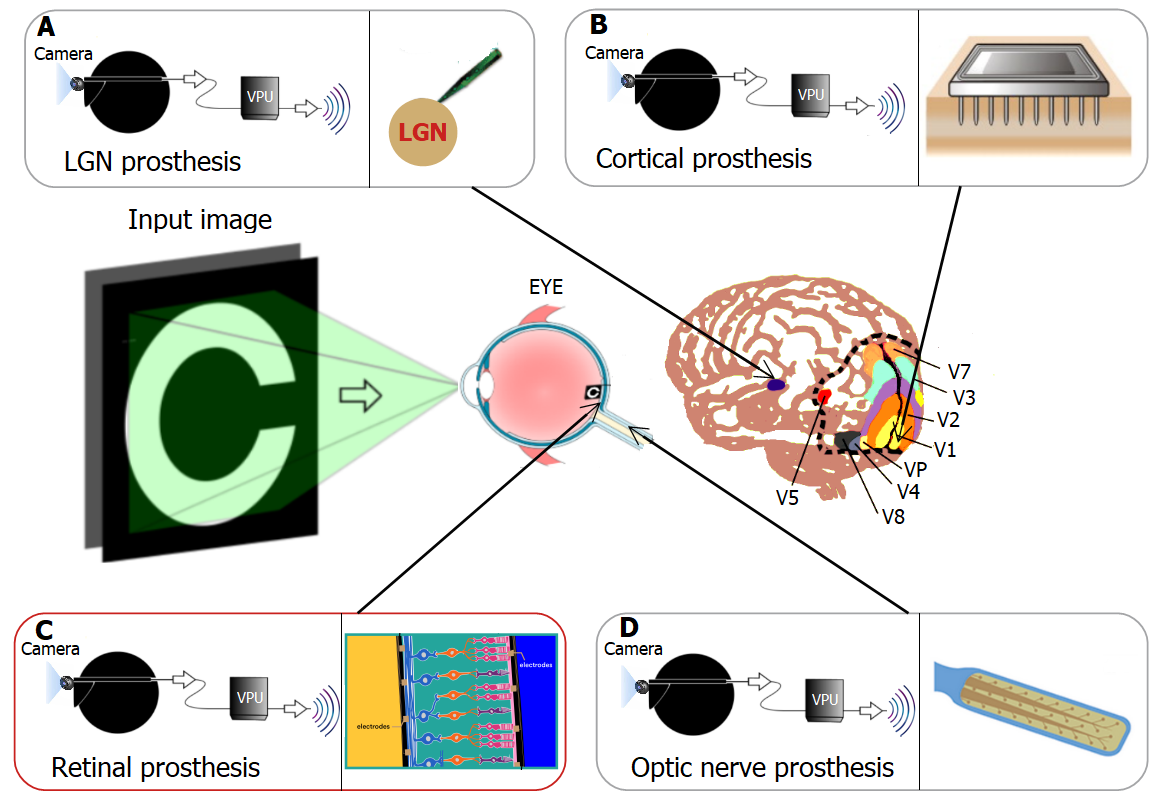}}
\caption{\textbf{Main approaches for the design of a visual prosthesis.} A) Schematic diagram of a LGN prosthesis. B) Cortical prosthesis. C) Retinal prosthesis. D) Optic nerve prosthesis. In general, all the approaches share a common set of components: a camera to capture images, generally mounted on glasses; a video processing unit (VPU) that transform the visual scene into patterns of electrical stimulation and transmits this information through a radio-frequency link to the implanted device, and an electrode array implanted at some level in the visual pathways which has to be located near the target neurons.}
\label{Overview}
\end{figure}

Visual prostheses are presently the most viable technology for the treatment of low vision and there are many types being proposed because of its potential for the development of various types of devices with existing technologies (see Figure~\ref{Overview}). The basic concept of a visual prosthesis is "electrically stimulating nerve tissues associated with vision (such as the retina) to help transmit electrical signals with visual information to the brain" \cite{farnum2020new}. Thus, several research groups are focusing their efforts on the development of new approaches for artificial vision based on electric stimulation of the retina \cite{da2016five,lorach2016retinal,stingl2017interim}, optic nerve \cite{duret2006object,lu2013electrical,gaillet2020spatially}, lateral geniculate nucleus \cite{vurro2014simulation,killian2016perceptual}, or visual cortex \cite{fernandez2005development,normann2009toward,kane2013electrical,normann2016clinical,fernandez2018development,niketeghad2019phosphene}, as can be seen in Figure~\ref{Overview}. All of these prosthetic devices work by exchanging information between the electronic devices and different types of neurons, and although most of them are still in development, they show promise of restoring vision in many forms of blindness.

At present, retinal prostheses are the most successful approach in this field \cite{da2016five,stingl2017interim}. In a retinal implant a multielectrode array is set up on the retinal surface and stimulates the retina from the top side with electrodes. All this current implantable system are designed with electrodes implanted in the body working together with several devices worn outside the body. Thus, a visual prosthesis incorporates an external video camera for image acquisition, an image processor converting the image to a suitable pattern of electrical stimulation, and finally the electrical stimulation array on the retina itself  \cite{chader2009artificial,dowling2009current,sachs2004retinal,weiland2005retinal}. In spite of reports showing retinal prostheses capable of helping some participants perform simple tasks of daily living, such as detecting lights, recognizing objects and even reading large letters, there are still physiological and technological limitations of the information received by implanted patients. The number of electrodes and implant size limit the maximum amount of information that can be provided by the stimulating array. This fact has restricted the degree of visual resolution (up to 1500 phosphenes) and dynamic range of the visual perception (8 grey levels) that can be delivered to the user. Besides, current systems such as retinal implants provide a field of view (FOV) of approximately $18^\circ$×$11^\circ$ in the retinal area, which correspond to the FOV subtended by the electrode implant on the retina. Moreover, the visual acuity of existing devices is very low, which means that crucial skills such as facial recognition or navigation in unknown environments are not yet possible. 

The visual acuity in prosthetic vision is limited by various factors from both engineering and physiological perspectives \cite{shim2020retinal}. One of the main causes of low visual acuity is the limited spatial resolution that can be achieved by electrical stimulation with existing visual implants. For example, the size of the electrodes in today's retinal implants is often much larger than the size of the neurons in the retina, and the number of electrodes is low \cite{tong2019improved}. However, it is not entirely true that visual perception will improve by just increasing the number of electrodes \cite{fernandez2020toward}. In addition, it has been shown that increased FOV is associated with a significant improvement in visual acuity \cite{ameri2009toward}. Several retinal prostheses have been tested in clinical trials \cite{rizzo2014argus,da2016five,humayun2012interim,stingl2015subretinal,zrenner2011subretinal} but although the results are very encouraging, still have to provide higher visual acuities to allow blind users to perform daily activities \cite{stronks2014functional,sahel2013acuboost}. These devices have been shown to restore vision up to a visual acuity of 1.8 logMAR and 1.44 logMAR, respectively. Regardless, the optimal number of the electrodes and FOV to provide adequate prosthetic vision is still an open question and an important design parameter needed to develop better implants.

In order to gain a better understanding of the potential benefits of low resolution visual prostheses we can use Simulated Prosthetic Vision (SPV). The SPV system is a standard procedure for non-invasive evaluation using normal vision subjects. Generally, in SPV, a low-resolution image of the view is presented on a computer screen \cite{sanchez2020semantic} or on a head-mounted display \cite{sanchez2020influence,chen2004effect,chen2005visual} glasses to a normally sighted user (see Figure~\ref{DataProcess}). It allows researchers to rigorously investigate the minimal requirements for a functional visual prosthesis and to explore which variables are important in the development of a visual prosthesis. However, the phosphenes perceived by people with visual prostheses are not yet fully understood. Realistic perception simulations with visual implants would be useful for the development and evaluation of future visual prosthetic systems. Avraham et al. \cite{avraham2021retinal} implemented a retinal prosthetic vision simulation, including temporal aspects such as persistence and perceptual fading of phosphenes and the electrode activation rate. Other previous works have focused on development of computational models to describe some of these distortions for a small number of behavioral observations in either space \cite{nanduri2012frequency} or time \cite{horsager2009predicting}. Beyeler et al. \cite{beyeler2017pulse2percept} go further and simulate spatial distortions, temporal nonlinearities, and spatio-temporal interactions reported across a wide range of conditions, devices, and patients.

A major outstanding challenge is predicting what people ‘see’ when they use their devices or how the implanted subjects will see with the future devices. Another challenge is addressing the narrow FOV found in most devices as well as the number and size of electrodes. Some SPV studies attempt to address this using computer monitors but this requires patients to scan the environment with head movements while trying to piece together the information, which is difficult to measure with static monitors. To address these challenges, SPV can be also assessed in controlled, real, or virtual environments. For instance, a SPV model in immersive virtual-reality (VR) allowing sighted subjects to act as implanted patients. In this setup, the visual input about to be rendered to a head-mounted display (HMD) mimics the external camera of a retinal implant. This input can come from the HMD’s camera or can be simulated in a virtual environment. This allows sighted subjects to ‘see’ through the eyes of a retinal prosthesis patient, taking into account their head. Some researchers have used the combination of SPV with a VR system for experimentation. Sanchez et al. \cite{sanchez2020influence} analyzed the influence of field of view with respect to resolution in visual prostheses through a study with a SPV setup using a VR system. Kasowski et al. \cite{kasowski2021towards} proposed to embed biologically realistic models of SPV in immersive VR so that sighted subjects can act as virtual patients in real-world tasks. Thorn et al. \cite{thorn2020virtual} implemented prosthetic vision in a VR environment in order to simulate the real-life experience of using a retinal prosthesis and investigated the interaction between the field of view and the pixel number. Moreover, prosthetic visual acuity for rectangular and hexagonal phosphene grids was also tested using a virtual reality simulation \cite{chen2005visual}. Similarly, Chen et al. \cite{chen2004effect} examined visual acuity of prosthetic vision under VR simulation measuring parameters such as filtering scheme, filter aperture and the phosphene matrix.


In this work, we take advantage of virtual-reality software paired with a portable head-mounted display and evaluated the performance of normally sighted participants under simulated prosthetic vision with variable field of view and number of pixels. In our system, the head-mounted display mimics the external camera of a visual implant subjects and allows simple experimentation in order to study the design parameters of future visual prostheses.  



\section{Methods}

We examined visual acuity on a stimuli recognition task using SPV through a VR system. The SPV system is a standard procedure for non-invasive evaluation using normal vision subjects. This methodology allows controlled evaluation of normally sighted subject response and task performance which is fundamental to know the way humans perceive and interpret phosphenized renderings. SPV also offers the advantage of adapting implant designs to improve the perceptual quality without involving implanted subjects.

\subsection{Participants}
Ten subjects with normal vision volunteered for the formal experiment. The subjects (four females and six males) were between 22 and 35 years old. Every subject used a computer daily. 

\subsubsection{Ethical statement}

The research process was conducted according to the ethical recommendations of the Declaration of Helsinki. The research protocol used for this study is non-invasive, purely observational, with absolutely no-risk for any participant. There was no personal data collection or treatment and all subjects were volunteers. Subjects gave their informed written consent after explanation of the purpose of the study and possible consequences. The consent allowed the abandonment of the study at any time. All data were analyzed anonymously. The experiment was approved by the Aragon Autonomous Community Research Ethics Committee (CEICA, see Ethical Statement for additional details).

\subsection{Simulated Prosthetic Vision (SPV)}

\begin{figure}[t!]
\centering
{\includegraphics[width = 6in]{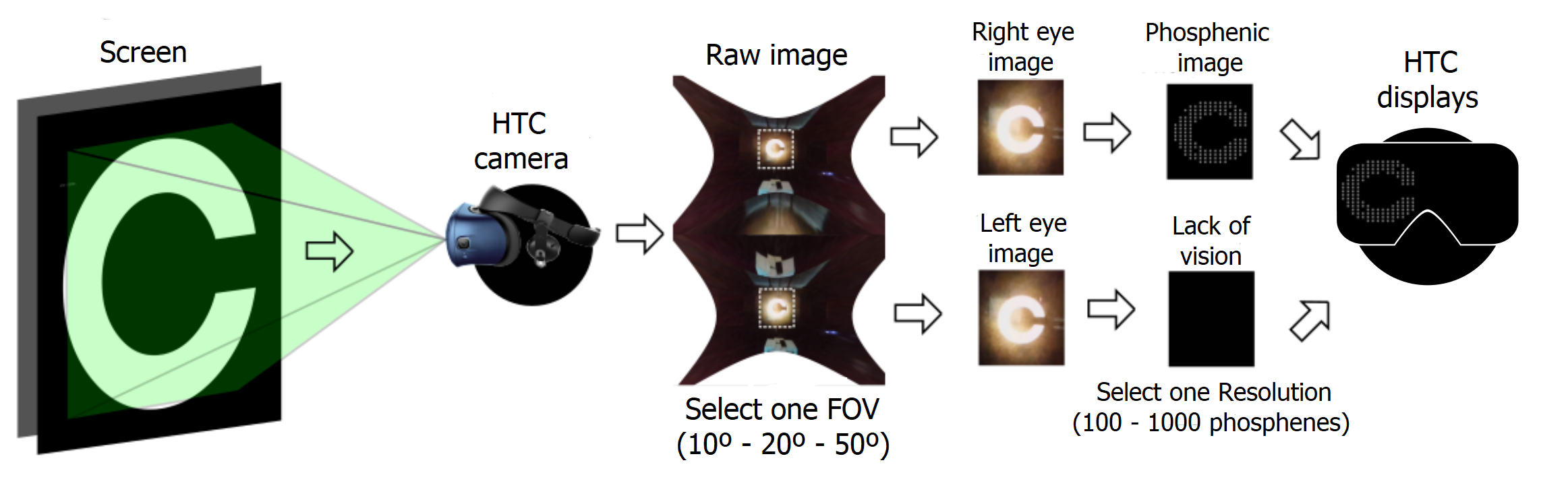}}
\caption{\textbf{Data process}. From the images obtained by the HTC cameras we cut the central area of the raw images to eliminate some kind of edge distortion. The selected area is projected on the two HTC displays. Finally, we convert the images into simulated phosphenes. We only project the phosphenic image to the right eye, simulating an implant in the right eye, and lack of vision on the left.}
\label{DataProcess}
\end{figure}

This section describes the SPV system including the hardware specifications, software components and phosphene generation.

\subsubsection{Hardware}

The experiment was conducted on an HTC VIVE PRO powered by a computer (Intel(R) Core(TM) i9-9900KF CPU 3.60GHz, NVIDIA GeForce RTX 2080 Ti). The VR system is composed by two lenses, two screens, SteamVR Tracking, G-sensor, gyroscope, proximity and Eye Comfort Setting (IPD). It contains dual AMOLED 3.5" diagonal screen with a resolution of 1440 x 1600 pixels per eye (2880 x 1600 pixels combined), covering a visual field of approximately 110 degrees. In our experiments we mostly use the central part of the display which remains undistorted. The representation with simulated phosphenes was displayed on the VR system worn by the participants as well as on the computer screen for the experimenter to check the progress. During the experiment, participants were seated in a backless chair allowing them to scan the entire scene with head rotation movements.

\subsubsection{Software}
The implementation was done in C++, using OpenVR for HTC VIVE Pro to connect with the VR system and OpenCV for image processing. Our software is compatible with the Windows operating system. Figure~\ref{DataProcess} shows the data process designed to generate the stimuli for the VR system. From the images obtained by the HTC cameras we cut the central area of the raw images to eliminate some kind of edge distortion. The selected area is projected on the two HTC displays. Finally, we convert the images into simulated phosphenes. We only project the phosphenic image to the right eye, simulating an implant in the right eye, and lack of vision on the left (see Figure~\ref{DataProcess}).

Our phosphene map configuration is similar to the framework of Sanchez et al. \cite{sanchez2020influence}. We approximate the phosphenes as circular dots with a Gaussian luminance profile —each phosphene has maximum intensity at the center and gradually decays to the periphery, following a Gaussian function–. The intensity of a phosphene is directly extracted from the intensity of the same region in the image. For our experiments, each phosphene has eight intensity levels. The size and brightness are directly proportional to the quantified sampled pixel intensities. The phosphene map is calculated and updated with respect to head orientation in real time. 

\subsection{Procedure}

\begin{figure}[t!]
\centering
\includegraphics[width = 4in]{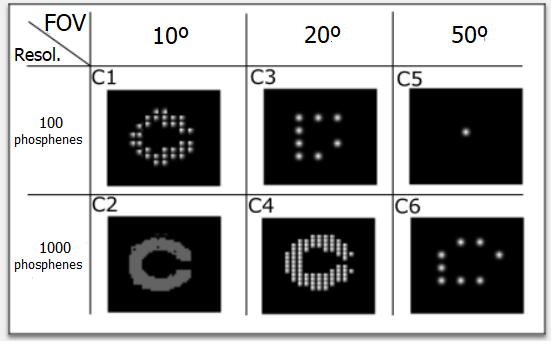}
\caption{\textbf{Stimuli conditions in the experiment.} The six possible stimulus conditions are depicted for the `Landolt-C orientation test'. `FOV-Resolution': C1: $10^\circ$-100 phosphenes, C2: $10^\circ$-1000 phosphenes, C3: $20^\circ$-100 phosphenes, C4: $20^\circ$-1000 phosphenes, C5: $50^\circ$-100 phosphenes and C6: $50^\circ$-1000 phosphenes.}
\label{Conditions}
\end{figure}

The experiment was conducted using a selection of stimulus from Balm \cite{bach2010basic} and Freiburg \cite{bach1996freiburg} tests, adapted to our SPV system, which are automated procedures for self-administered measurement of visual acuity. The images were presented to the subjects using different stimuli conditions based on two resolutions (100 and 1000 phosphenes) and three FOVs (10, 20 and 50 degrees), as can be seen in Figure~\ref{Conditions}. We selected these particular resolutions and FOVs based on current visual prostheses \cite{bloch2019advances,rizzo2014argus,da2016five,humayun2012interim,stingl2015subretinal,zrenner2011subretinal}, although our VR platform allows to quickly change those parameters.

\begin{figure}[t!]
\centering
\includegraphics[width = 6in]{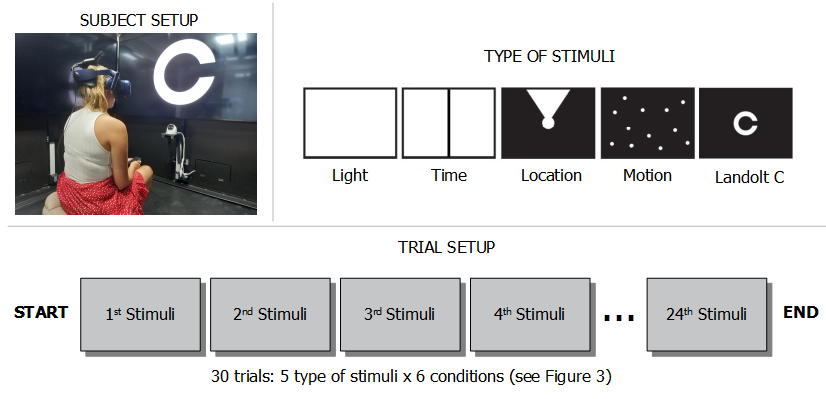}
\caption{\textbf{Subject and trial setup.} Subjects view through the HMD what is shown in the picture on the monitor. Subjects scanned the underlying picture with their head motion. They used the joystick to indicate the orientation of the stimuli for subject response. For the experiment, we used five stimuli from Balm \cite{bach2010basic} and Freiburg \cite{bach1996freiburg} tests: Light, Time, Location, Motion, and Landolt C. Each test consisted of 24 stimuli. The stimuli in each test were randomly selected.}
\label{trial_setup}
\end{figure}

The participants performed five tests based on different types of stimuli, described as `light perception', `time recognition', `light location', `motion perception' and `Landolt-C orientation' (see Figure~\ref{trial_setup}). The first stimulus corresponded to the `light perception' and is the simplest stimulus of the experiment that tests the basic perception of light. The subjects' task was to decide whether they see the light appear after the warning tone or not. The second stimulus corresponded to `time resolution', which assesses one basic aspect of time resolution—namely, whether one or two flashes occur after an indicator beep. The third stimulus corresponded to the `light location' that tests the projection of light. A light disc appeared that the subject must center in the limited visual field. After a pre-set delay, simultaneously with a warning tone, a wedge appeared directed up, down, right, or left from the fixation disc. The fourth stimulus corresponded to `motion perception'. A random hexagonal pattern of light and dark elements appeared. After an acoustic signal, it began to move in one of four directions (up-down-right-left). The subject indicated the motion’s direction. The last stimulus corresponded to the `Landolt-C orientation' test. Subjects had to indicate the orientation in one of four directions (up-down-right-left) of the gap in the Landolt-C, which is a standard international symbol for testing visual acuity. In all tests subjects responded via corresponding joystick positions. The number of trials was set to 24 for all the tests. 

Subjects sat in a chair that was adjustable in height and were instructed to look straight at the middle of the screen. A HMD was used in order to immerse the subject into a virtual-reality environment. Subjects view through the HMD what was shown in the picture on the computer monitor. Subjects scanned the underlying picture with their head motion. The image was projected to the right eye only, simulating an implant in the right eye, and lack of vision on the left (see Figure~\ref{DataProcess}). The subjects were shielded from as much ambient light as possible and lights were turned off in the room. The computer generated graphics were simulated to be stationary in space; head motion corresponded to scanning an image, sampling with the phosphenized view at different locations of the visual space. 

Before the start of each trial, the subjects were informed about that stimulus functions and the valid choices and they carried out a training test with the 5 stimuli with normal vision to become familiar with the tasks and the environment. To avoid errors during the experiment, we trained the subjects to enter their responses via a keypad with four keys (left, right, top, and bottom). The keypad was a commercial USB-connected entry pad. The four keys coded the test response: light, left key; no light, right key; one flash, left key; two flashes, right key. For the `light location', `motion perception' and `Landolt-C orientation' tests, the keys corresponded to the observed direction. They were encouraged to respond within the time limit, even when uncertain. No head or eye tracking was used during the testing to account for fixation.

\subsection{Statistical analysis}

Data were analyzed using two-way ANOVA and post hoc-test with Tukey’s method to evaluate simultaneously the effect of the two grouping variables (resolution and FOV) on the response variables performance and reaction time with $p= 0.05$, $*<0.05$ , $**<0.01$, $***<0.001$ and $ns$ not significant.

\section{Results}

The results are summarized from Figure~\ref{fig5} to Figure~\ref{VAcuity}. All figures show box-plots of data distribution with 25, 50 and 75\textit{th} quartiles for performance and reaction time (mean $\pm$ standard deviation) for aggregated data from all subjects. The performance (in percentage) is defined as number of correct responses. The reaction time (in seconds) is the time from the subject’s first response. We also performed a test to determine if the mean difference between specific pairs of conditions are statistically significant using Tukey’s method with a significant level $ \alpha = 0.05$.

\subsection{Light perception}

\begin{figure}[t!]
\centering
\subfigure[]{\includegraphics[width = 3in]{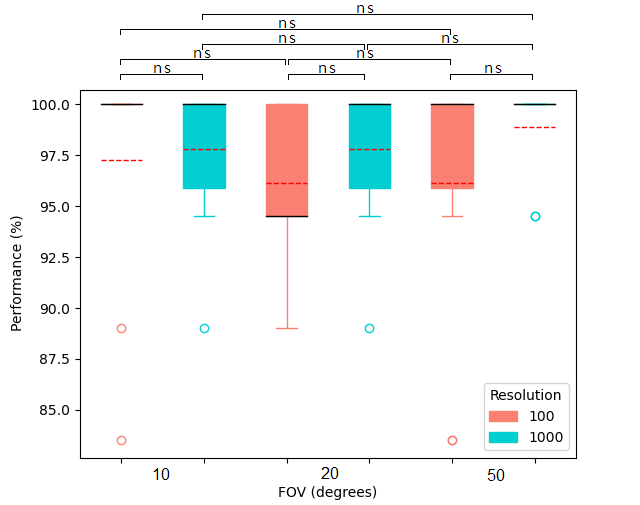}\label{Li_Bar_Perf}}
\subfigure[]{\includegraphics[width = 3in]{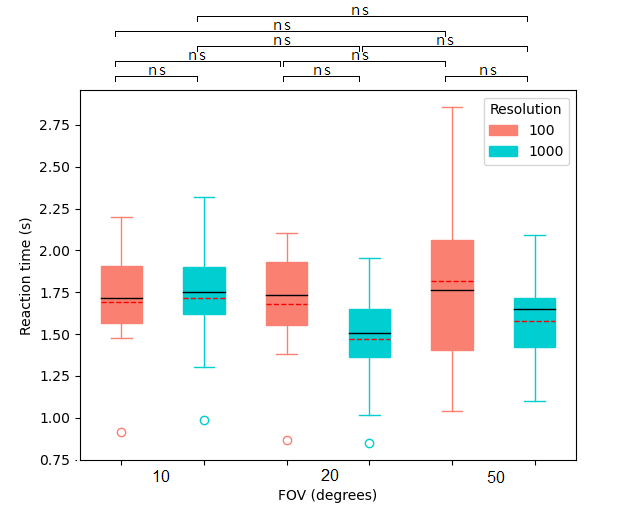}\label{Li_Bar_Time}}
\caption{\textbf{Light perception.} Performance and reaction time for the `light perception' test. \subref{Ti_Bar_Perf} Bar-plot for performance. \subref{Ti_Bar_Time} Bar-plot for time.}
\label{fig5}
\end{figure}

Figure~\ref{fig5} shows the performance and reaction time for the `light perception' test. This test is the simplest of the five tests carried out in the experiment. All subjects were able to perceive light or not in almost all trials. All the conditions obtain high values above 95\%, as can be seen in Figure~\ref{Li_Bar_Perf}. There was no significant difference between the stimulus conditions. The reaction time for all conditions are similar close to 1.6s (see Figure~\ref{Li_Bar_Time}). The lowest reaction time was $1.47 \pm 0.34$, corresponding to the condition of $20^\circ$ and 1000 phosphenes. For the resolution of 100, no significant difference was found for 10-20 FOVs (p=0.9964), 20-50 FOVs (p=0.7723) and 10-50 FOVs (p=0.8174). For the resolution of 1000, nor significant difference was found for 10-20 FOVs (p=0.2497), 20-50 FOVs (p=0.7680) and 10-50 FOVs (p=0.6223). There was no significant difference between the two resolutions ($p>0.05$). Some outliers can be observed in both graphs corresponding to moments in which the subjects were distracted during the task.



\subsection{Time resolution}

\begin{figure}[t!]
\centering
\subfigure[]{\includegraphics[width = 3in]{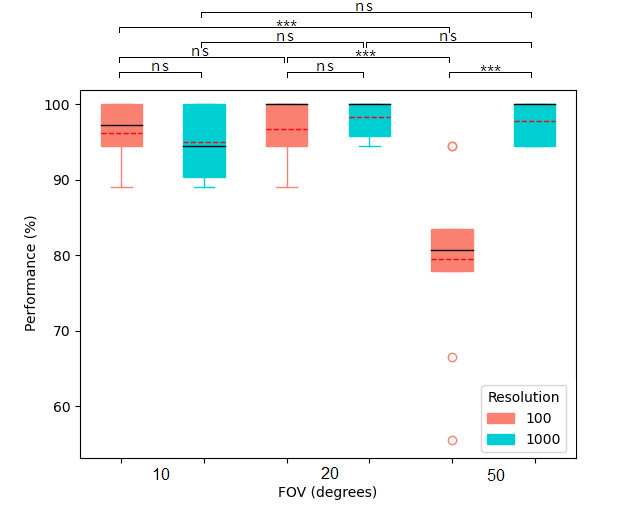}\label{Ti_Bar_Perf}}
\subfigure[]{\includegraphics[width = 3in]{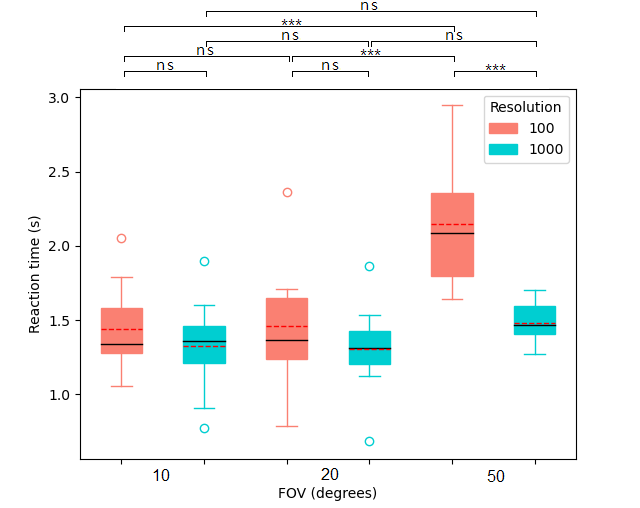}\label{Ti_Bar_Time}}
\caption{\textbf{Time resolution}. Performance and reaction time for the `time resolution' test. \subref{Ti_Bar_Perf} Bar-plot for performance. \subref{Ti_Bar_Time} Bar-plot for time.}
\label{fig6}
\end{figure}

Figure~\ref{fig6} shows the performance and reaction time for the `time resolution' test. For the resolution of 100 phosphenes, the average performance is $96.15 \pm 4.53$, $96.70 \pm 4.64$ and $79.55 \pm 11.78$ for 10, 20 and 50 degrees respectively (see Figure~\ref{Ti_Bar_Perf}). No significant difference was found for 10-20 FOVs (p=0.9863). However, significant difference was found for 10-50 and 20-50 FOVs. For the resolution of 1000, the average performance is $95.05 \pm 4.82$, $98.35 \pm 2.66$ and $97.80 \pm 2.84$ for 10, 20 and 50 degrees respectively. No significant difference was found for 10-20 FOVs (p=0.1163), 20-50 FOVs (p=0.2160) and 10-50 FOVs (p=0.9369).

Figure~\ref{Ti_Bar_Time} shows the reaction time for the `time resolution' test. For the resolution of 100 phosphenes, the average reaction time is $1.44 \pm 0.31$, $1.46 \pm 0.42$ and $2.15 \pm 0.43$ for 10, 20 and 50 degrees respectively. No significant difference was found for 10-20 FOVs (p=0.9947). For the resolution of 1000, the average reaction time is $1.33 \pm 0.32$, $1.31 \pm 0.30$ and $1.48 \pm 0.14$ for 10, 20 and 50 degrees respectively. No significant difference was found for 10-20 FOVs (p=0.9877), 20-50 FOVs (p=0.3160) and 10-50 FOVs (p=0.3909).

\subsection{Light location} 

\begin{figure}[t!]
\centering
\subfigure[]{\includegraphics[width = 3in]{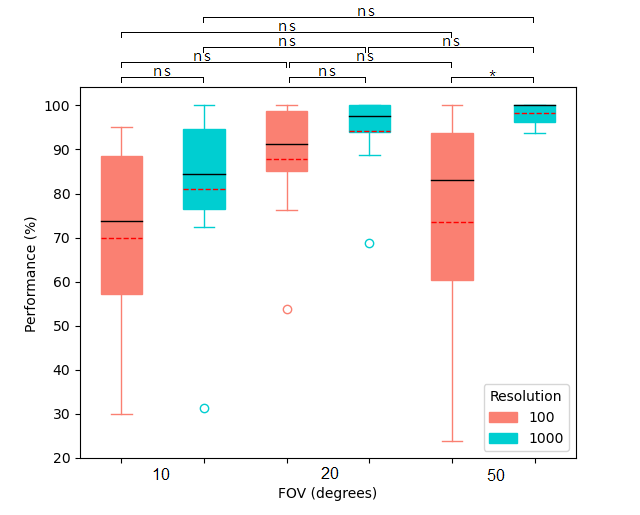}\label{Lo_Bar_Perf}}
\subfigure[]{\includegraphics[width = 3in]{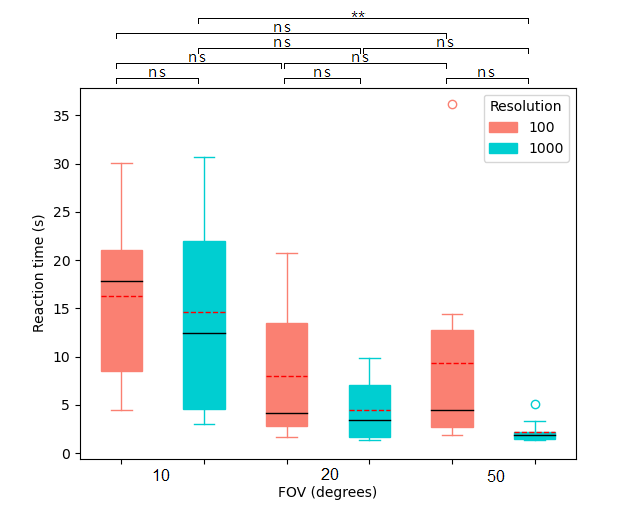}\label{Lo_Bar_Time}}
\caption{\textbf{Light location}. Performance and reaction time for the `light location' test. \subref{Ti_Bar_Perf} Bar-plot for performance. \subref{Ti_Bar_Time} Bar-plot for time.}
\label{fig7}
\end{figure}

Figure~\ref{fig7} shows the performance and reaction time for the `light location' test. The performance increases as the FOV and the resolution increases (see Figure~\ref{Lo_Bar_Perf}). For the resolution of 100 phosphenes the average performance is $70.00 \pm 21.79$, $87.75 \pm 14.37$ and $73.63 \pm 26.15$ for 10, 20 and 50 degrees respectively. No significant difference was found for 10-20 FOVs (p=0.1697), 20-50 FOVs (p=0.3158) and 10-50 FOVs (p=0.9237). For the resolution of 1000 phosphenes the average performance is $81.00 \pm 19.89$, $94.13 \pm 9.70$ and $98.38 \pm 2.64$ for 10, 20 and 50 degrees respectively. No significant difference was found for 20-50 FOVs (p=0.7512). 

Figure~\ref{Lo_Bar_Time} shows the reaction time for the three FOVs and two resolutions. For the resolution of 100 phosphenes the average reaction time is $16.25 \pm 8.63$, $8.00 \pm 6.82$ and $9.40 \pm 10.56$ for 10, 20 and 50 degrees respectively. No significant difference was found for 10-20 FOVs (p=0.1099), 20-50 FOVs (p=0.9335) and 10-50 FOVs (p=0.2089). For the resolution of 1000 phosphenes the average reaction time is $14.62 \pm 10.98$, $4.47 \pm 3.27$ and $2.24 \pm 1.17$ for 10, 20 and 50 degrees respectively. No significant difference was found for 20-50 FOVs (p=0.6704). Comparing the performance for the same FOV, the reaction time decreases with increasing number of phosphenes.

\subsection{Motion perception}

\begin{figure}[t!]
\centering
\subfigure[]{\includegraphics[width = 3in]{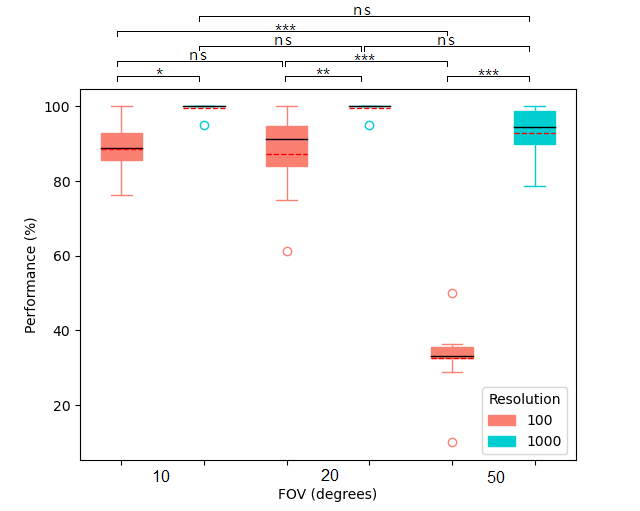}\label{Mo_Bar_Perf}}
\subfigure[]{\includegraphics[width = 3in]{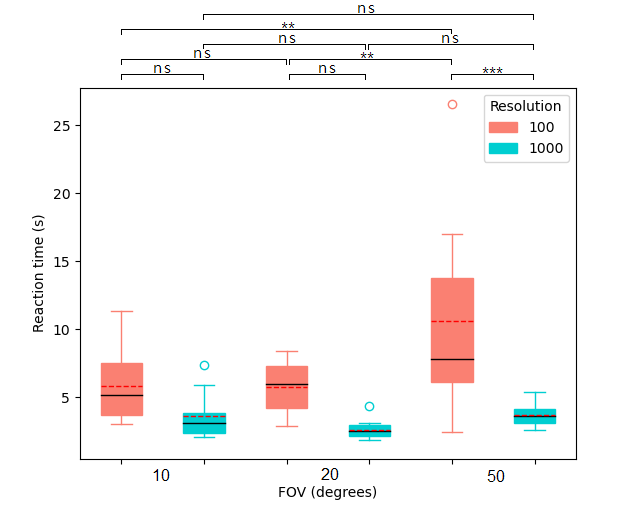}\label{Mo_Bar_Time}}
\caption{\textbf{Motion perception}. Performance and reaction time for the `motion perception' test. \subref{Ti_Bar_Perf} Bar-plot for performance. \subref{Ti_Bar_Time} Bar-plot for time.}
\label{fig8}
\end{figure}

Figure~\ref{fig8} shows the performance and reaction time for the `motion perception' test. For the same resolution, the performance decreases as the FOV increases (see Figure~\ref{Mo_Bar_Perf}). For the resolution of 100 phosphenes the average performance is $88.63 \pm 6.57$, $87.38 \pm 11.64$ and $32.63 \pm 9.76$ for 10, 20 and 50 degrees respectively. No significant difference was found for 10-20 FOVs (p=0.9540). For the resolution of 1000 phosphenes the average performance is $99.50 \pm 1.58$, $99.50 \pm 1.58$ and $92.88 \pm 7.17$ for 10, 20 and 50 degrees respectively. No significant difference was found for 10-20 FOVs (p=1.000). 

Figure~\ref{Mo_Bar_Time} shows the reaction time for the `motion perception' test. For the same resolution, the reaction time increases as the FOV increases. For the resolution of 100 phosphenes the average reaction time is $5.84 \pm 2.73$, $5.80 \pm 1.94$ and $10.59 \pm 7.13$ for 10, 20 and 50 degrees respectively. No significant difference was found for 10-20 FOVs (p=0.9998), 20-50 FOVs (p=0.0647) and 10-50 FOVs (p=0.0675). For the resolution of 1000 phosphenes the average reaction time is $3.64 \pm 1.72$, $2.64 \pm 0.75$ and $3.71 \pm 0.85$ for 10, 20 and 50 degrees respectively. No significant difference was found for 10-20 FOVs (p=0.1591), 20-50 FOVs (p=0.1265) and 10-50 FOVs (p=0.9915). Comparing the performance for the same FOV, the reaction time decreases with increasing number of phosphenes.

\subsection{Landolt-C orientation}

\begin{figure}[t!]
\centering
\subfigure[]{\includegraphics[width = 3in]{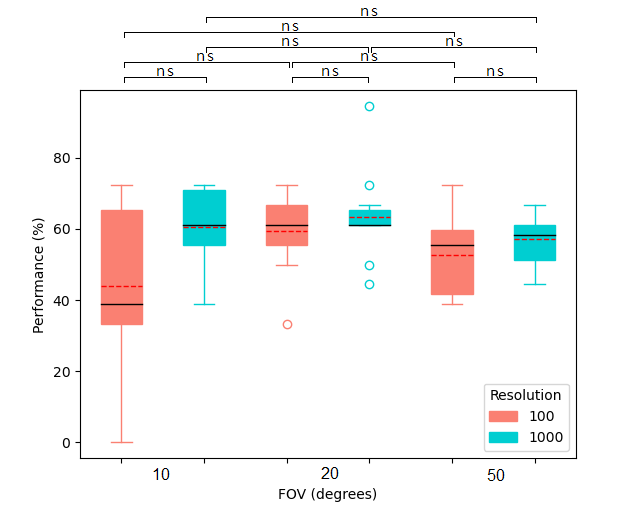}\label{fract_Bar_Perf}}
\subfigure[]{\includegraphics[width = 3in]{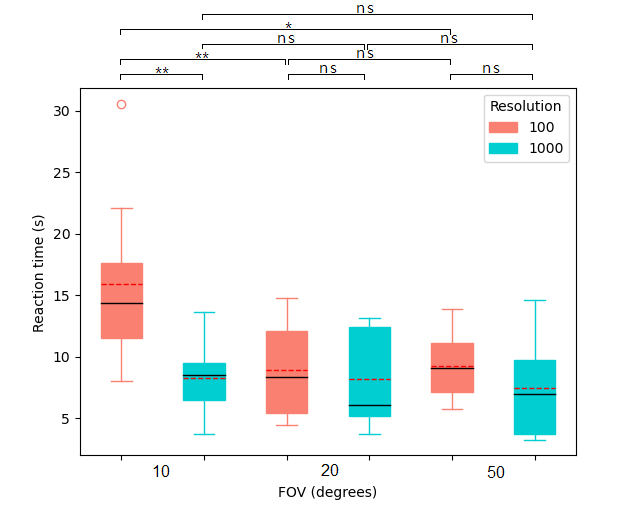}\label{fract_Bar_time}}
\caption{\textbf{Landolt-C orientation}. Performance and reaction time for the `Landolt-C orientation' test. \subref{Ti_Bar_Perf} Bar-plot for performance. \subref{Ti_Bar_Time} Bar-plot for time. }
\label{fig9}
\end{figure}

Figure~\ref{fig9} shows the performance and reaction time for the `Landolt-C orientation' test. For the resolution of 100 phosphenes the average performance is $43.89 \pm 23.78$, $59.44 \pm 12.02$ and $52.78 \pm 11.19$ for 10, 20 and 50 degrees respectively (see Figure~\ref{fract_Bar_Perf}). No significant difference was found for 10-20 FOVs (p=0.1141), 20-50 FOVs (p=0.6499) and 10-50 FOVs (p=0.4732). For the resolution of 1000 phosphenes the average performance is $60.65 \pm 10.94$, $63.33 \pm 13.41$ and $957.22? \pm 7.43$ for 10, 20 and 50 degrees respectively. No significant difference was found for 10-20 FOVs (p=0.8366), 20-50 FOVs (p=0.4314) and 10-50 FOVs (p=0.7738).

For the resolution of 100 phosphenes the average reaction time is $15.92 \pm 6.53$, $8.92 \pm 3.93$ and $9.24 \pm 2.70$ for 10, 20 and 50 degrees, respectively. No significant difference was found for 20-50 FOVs (p=0.9874) (see Figure~\ref{fract_Bar_time}). For the resolution of 1000 phosphenes the average reaction time is $8.29 \pm 2.91$, $8.17 \pm 3.97$ and $7.46 \pm 4.16$ for 10, 20 and 50 degrees respectively. No significant difference was found for 10-20 FOVs (p=0.9970), 20-50 FOVs (p=0.9041) and 10-50 FOVs (p=0.8711).

Figure~\ref{VAcuity} shows the values of visual acuity (in logMAR) for the `Landolt-C orientation' test obtained for each condition. Comparing the same FOV, we obtain higher visual acuity for 1000 resolution than for 100. However, the most significant difference between the two resolutions was found for the 10 FOV. The best visual acuity is obtained for the condition of $20^\circ$ and 1000 resolution, with a visual acuity value of 1.3 logMAR. This visual acuity is considered the limit of blindness.

\section{Discussion}

During the past decade, much effort has been put into developing visual prosthesis devices that help restore vision in blind people \cite{da2016five,lorach2016retinal,stingl2017interim,duret2006object,lu2013electrical,gaillet2020spatially,vurro2014simulation,killian2016perceptual,fernandez2005development,normann2009toward,kane2013electrical,normann2016clinical,fernandez2018development,niketeghad2019phosphene}. However, much less has been spent on finding acceptable procedures to assess the functionality of different visual implant technologies and maximising the benefits of artificial vision.

One of the most common methods used in order to gain a better understanding of the potential benefits of low resolution visual prostheses is through simulated prosthetic vision (SPV). Generally, in a SPV experiment, a real-time, low-resolution image of the view is presented on a HMD to a normally sighted subject. The image of the scene is captured by a head mounted camera, digitalized by a computer, and a sub-sampled low resolution image is presented on the HMDs. In this way, a variety of tasks have been evaluated using SPV and encouraging performance results were reported on reading speed \cite{ho2019performance,paraskevoudi2021full,abraham2019active,mandel2019reading}, navigation \cite{sanchez2021augmented,lo2021navigation,vergnieux2017simplification}, object recognition \cite{sanchez2020semantic,li2018image}, hand–eye coordination \cite{titchener2018gaze} and face recognition \cite{irons2017face,ho2019performance}, among others. However, visual acuity tests are the principle quantitative measures used to assess the efficacy and cost effectiveness of procedures designed to improve or restore vision \cite{rosenfeld2006ranibizumab,kobelt2002cost}. In a visual acuity test a patient is required to report the identity of different patterns presented in various sizes (spatial frequency), and the resulting visual acuity is defined by the smallest shape that can be correctly identified by the observer. Visual acuities of 0.5 logMAR and 1.0 logMAR are considered moderate and severe visual impairment, respectively. A visual acuity greater than 1.3 logMAR is considered total blindness. The theoretical visual acuity achievable by present-day retinal implants such as Argus II with a FOV of approximately 20$^{\circ}$ is 2.5 logMAR (20/6325), and the highest acuity is shown to be 1.8 logMAR (20/1262) \cite{humayun2012interim,ho2015long} in several basic tasks, such as those concerning grating visual acuity, square localization, and movement detection \cite{zhou2013argus,ho2015long,humayun2009preliminary,ahuja2011blind}. Clinical trials for the Alpha-IMS assessed the visual acuity and object recognition of the restored vision in the context of daily living and mobility \cite{stingl2015subretinal}. Three patients were able to read letters with a visual angle between 5$^{\circ}$ and 10$^{\circ}$, demonstrating the highest visual acuity of 1.44 logMAR in the Landolt-C test. The visual acuity achieved by the Alpha-IMS is not significantly higher than that of the Argus II, given its 25-fold higher number of stimulating channels. These results may imply that the patterned stimulation generated from all the 1500 individual channels of the Alpha-IMS could not be perfectly discriminated by the retinal cells, lowering the perceived spatial.


\begin{figure}[t!]
\centering
{\includegraphics[width = 3in]{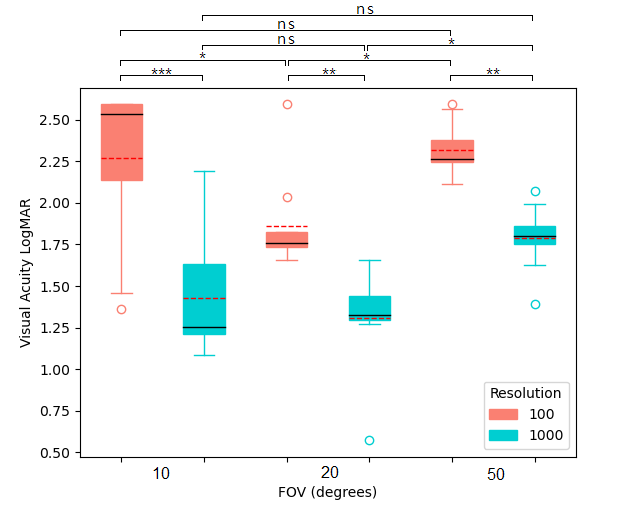}}
\caption{\textbf{Visual acuity for `Landolt-C orientation' test.} Values of visual acuity (in logMAR) for the `Landolt-C orientation' test obtained for each condition. }
\label{VAcuity}
\end{figure}

Visual acuity has already been used in the past both in clinical trials of visual neuroprosthesis \cite{humayun1996visual,dobelle2000artificial} but also in simulation experiments of prosthetic vision \cite{cha1992simulation,chen2004effect,hayes2003visually}. The SPV software samples the image to match the resolution of the retinal prosthesis devices. By using SPV, researchers evaluated whether and under what conditions a measured visual acuity level is a true indication that the visual prosthesis provides a patterned image. Visual acuity is classically measured by optotypes such as letters, numbers or Landolt C-rings. Just as the interest in developing a visual prosthesis intensified in recent years, some researchers published their SPV study findings regarding the number of phosphenes required for comparable-to-normal visual acuity. In 2004, Chen et al. \cite{chen2004effect} examined visual acuity under virtual-reality (VR) in SPV using different filtering schemes. The best mean score recorded by the subjects was 1.55 logMAR. Later, they tested visual acuity for both rectangular and hexagonal phosphene grids using the Freiburg test \cite{bach1996freiburg}. The visual acuity scores ranged from 1.45 to 1.80 logMAR depending on subject. Similarly, Cha et al. \cite{cha1992simulation} measured visual acuity as a function of the number of pixels and their spacing. They concluded that 625 electrodes implanted in a 1x1 cm area near the foveal representation of the visual cortex should produce a phosphene image with a visual acuity of approximately 20/30, 0.17 logMAR. Hayes et al. \cite{hayes2003visually} simulate three retinal implants and test the functionality of this vision with four-choice orientation discrimination of a Sloan letter E. Subjects were found to have visual acuities of 1.96, 1.82, and 1.32 logMAR with the 4x4, 6x10, and 16x16 electrode arrays, respectively. 

\begin{table}
\caption{\label{tab:table1}\textbf{ Pixel density on the phosphenic image.} Amount of pixels needed to form one phosphene on each of the conditions.}
\centering
\begin{tabular}{ccc}
\br
Resolution            & \begin{tabular}[c]{@{}c@{}}Field of view\\ (degrees)\end{tabular} & \begin{tabular}[c]{@{}c@{}}Pixel density\\ (Pixels/Phos)\end{tabular} \\
\br
\multirow{3}{*}{100}  & 10                                                                & 9                                                                         \\ 
                      & 20                                                                & 36                                                                        \\ 
                      & 50                                                                & 361                                                                       \\ 
\mr
\multirow{3}{*}{1000} & 10                                                                & 1                                                                         \\  
                      & 20                                                                & 4                                                                         \\
                      & 50                                                                & 36                                                                        \\ 
\br
\end{tabular}
\end{table}
\normalsize

We found that the recognition of the different stimuli is well achieved with low resolution and restricted FOV. As can be seen in Figures~\ref{fig5}, \ref{fig6}, \ref{fig7}, \ref{fig8} and \ref{fig9}, for almost all tests a significant improvement in task performance was obtained for a 20 FOV and a resolution of 1000. Besides, participants took less time to recognize the stimuli with this condition. Generally, participants took less time to recognize stimuli with the 20 FOV than the 50 FOV. This seems counterintuitive since with a narrower FOV the global reference of the image is lost. However, the narrower the FOV, the higher the angular resolution and therefore the greater the image detail (higher frequencies). Contrary, to generate one phosphene in the largest FOV, a larger number of pixels is averaged and therefore more information on image details is lost (see Table~\ref{tab:table1}). Thus, the widest FOV allows to cover the widest area of the image but it only allows to see the gist of the image (low spatial frequency). On the other hand, the higher visual acuity was also obtained with the condition of 20 FOV and 1000 resolution (see Figure~\ref{VAcuity}). For this condition, subjects obtained a visual acuity of 1.3 logMAR. If the number of phosphenes is reduced to 100 for the 20 FOV condition, visual acuity decreases to 1.86 logMAR. We can compare this visual acuity value with those obtained by some studies with Argus II using similar conditions of FOV and resolution \cite{humayun2012interim,ho2015long}. Furthermore, for the experimental condition of 10$^{\circ}$ FOV and 1000 resolution which can be compared to the Alpha-IMS implant \cite{stingl2015subretinal}, subjects obtained a visual acuity of 1.43 logMAR. 

However, visual acuity is not the only important parameter nor the spatial details of visual scenes. There are many other relevant aspects in visual scenes such as shape, color and movement that would allow the extraction of complex information, for example identifying human faces, from relatively poor-quality images by using specific cues and multiple visual features \cite{sinha2002recognizing} or obstacle detection from depth information and motion cues to facilitate the safe movement of the user in complex or unfamiliar environments \cite{perez2017depth}. This suggests that besides image resolution, we should try to pay attention to other relevant visual attributes such as receptive field size, localization, orientation, or movement \cite{fernandez2020toward}. In addition, depending on the subjects, there is one need or another. For example, some people focus more on identifying objects or people, while others prefer orientation and mobility. The key issue is to encode and send useful information that can be translated into functional gains for activities of daily living. Furthermore, it has been observed that there may be subtle differences in perceived visual field or encoding between subjects. Therefore, future advanced systems for interacting with the brain of people with low vision should allow the customization of functions to meet the needs of each subject.

\section{Conclusions}

Visual acuity tests are the main quantitative measures used to evaluate the effectiveness and cost-effectiveness of procedures designed to improve or restore vision. However, finding acceptable procedures for evaluating the functionality of visual implant technologies and maximizing the benefits of prosthetic vision is still under study. The present work constitutes a first essential step towards immersive virtual-reality simulations of prosthetic vision which has the potential to accelerate the prototyping of new devices. Via a head-mounted display, subjects were afforded simulated prosthetic vision (phosphene images) and required to recognise different stimuli normally used to measure visual acuity. Of all conditions tested, a FOV of 20$^{\circ}$ and 1000 phosphenes of resolution proved optimal, with higher visual acuity of 1.3 logMAR. Our simulated prosthetic vision system allows simple experimentation in order to study the design parameters of future visual prostheses. This work is a step toward the design of more effective electrode arrays that we hope will benefit the blind through neuroprosthesis.

\section*{Acknowledgments}

This work was supported by project RTI2018-096903-B-I00 (MINECO/FEDER, UE) and BES-2016-078426 (MINECO). The authors thank Violeta Estepa Ramos for collaborating in the development of the simulation environment.

\section*{References}

\bibliographystyle{iopart-num}
\bibliography{mybibfile}
\end{document}